\documentclass[runningheads]{llncs}
\usepackage[T1]{fontenc}
\usepackage{graphicx}
\begin{document}

{
    \renewcommand{\thefootnote}{\fnsymbol{footnote}}
    \footnotetext{\textit{Proc. of the Main Track of 22nd International Conference on Practical Applications of Agents and Multi-Agent Systems, 26th-28th June, 2024, https://www.paams.net/.}}
}

\title{Reinforcement Learning Enabled Peer-to-Peer Energy Trading for Dairy Farms}
%
%
\author{Mian Ibad Ali Shah \and
Enda Barrett \and
Karl Mason}
\authorrunning{Shah et al.}
%
\institute{School of Computer Science, University of Galway; Galway, Ireland, H91 FYH2\\
\email{\{m.shah7,enda.barrett,karl.mason\}@universityofgalway.ie }}
\maketitle              
\begin{abstract}

Farm businesses are increasingly adopting renewables to enhance energy efficiency and reduce reliance on fossil fuels and the grid. This shift aims to decrease dairy farms' dependence on traditional electricity grids by enabling the sale of surplus renewable energy in Peer-to-Peer markets. However, the dynamic nature of farm communities poses challenges, requiring specialized algorithms for P2P energy trading. To address this, the Multi-Agent Peer-to-Peer Dairy Farm Energy Simulator (MAPDES) has been developed, providing a platform to experiment with Reinforcement Learning techniques. The simulations demonstrate significant cost savings, including a 43\% reduction in electricity expenses, a 42\% decrease in peak demand, and a 1.91\% increase in energy sales compared to baseline scenarios lacking peer-to-peer energy trading or renewable energy sources.

\end{abstract}
\section{Introduction}

In accordance with the European climate law, EU member states are mandated to reduce greenhouse gas emissions by a minimum of 55\% by 2030, with the ultimate aim of achieving climate neutrality within the EU by 2050 \cite{eurcomclimate}. However, when considering dairy farming, particularly in countries like Ireland where a substantial portion of the economy relies heavily on dairy product exports, milk production involves significant energy consumption, giving rise to concerns regarding carbon dioxide ($CO_2$) emissions. In Ireland, dairy cows consume an average of 350 kilowatt-hours (kWh) of electricity annually \cite{upton2013energy}. Recent increases in energy prices have raised electricity expenses for farms, especially during peak utility grid hours \cite{upton2015investment}. Addressing this challenge requires the development of an artificial intelligence (AI) system to ensure the future sustainability of the dairy industry and aimed at reducing carbon emissions and peak grid demand, offering both environmental and financial benefits for dairy farms.

Multi-agent systems (MAS) consist of autonomous agents collaborating to achieve common goals, making decisions with limited information and communicating with immediate neighbours \cite{adjerid2020multi}. MAS shows promise in addressing microgrid challenges by providing scalable, decentralized control \cite{elena2022multi}. Performance models optimize profit for prosumers and meet end-user needs in energy-sharing regions (ESRs). Peer-to-peer (P2P) energy trading involves sharing energy within microgrids before trading with retailers \cite{zhou2018evaluation}.

P2P energy trading is still in its early stages of development but researchers have explored different approaches using centralized, decentralized, and distributed trading markets. Centralized markets face computational burdens and privacy concerns, while decentralized models offer direct P2P transactions \cite{umer2021novel}. Distributed approaches combine the benefits of both, and often use auction-based mechanisms like the Double Auction (DA) market to clear the energy trading market \cite{qiu2022mean}. 

\subsection{Centralized Trading}

In centralized trading, Qiu et al. proposed the PS-MADDPG algorithm, addressing prosumer heterogeneity to enhance energy trading by allowing agents in the same cluster to share experiences and learning policies \cite{qiu2021scalable}. Zhang et al. developed a MAS involving energy suppliers, employing non-cooperative games to achieve dynamic pricing and reduce peak demand \cite{zhang2019p2p}. Yang et al. introduced a multi-energy sharing strategy optimizing energy trading for better fairness using linear programming \cite{yang2022three}.

\subsection{Decentralized Trading}

Some researchers have pursued decentralized approaches. Guimaraes et al. utilized an ABM model to examine the economic benefits of P2P trading \cite{guimaraes2021agent}. Zhou et al. proposed a decentralized multi-agent model employing Q learning, Maulti-agent Reinforcement Learning (MARL), and Mid-Market Rate (MMR) to maximize revenue for each participant agent \cite{zhou2020decentralized}. Elkazaz et al. presented a hierarchical, decentralized approach considering factors like battery storage and PV size to examine their effect on the financial benefit for the community \cite{elkazaz2021hierarchical}. Lianyi Zhu investigated market-based and price-based optimal trading mechanisms using blockchain for maximized profits in a microgrid \cite{zhu2022market}. Khorasany et al. proposed decentralized P2P energy trading systems prioritizing data protection and financial benefits for participants \cite{khorasany2019decentralized}. 

\subsection{Distributed Trading}

Our focus lies in the distributed approach. Distributed markets require less information from peers and offer scalability and greater autonomy to customers \cite{khorasany2018market}. Shah et al. discussed the features of distributed P2P energy-sharing markets, blending centralized and decentralized aspects, enhancing customer privacy and autonomy \cite{shah2024multi}. Various studies categorized under distributed systems aimed at achieving financial benefit, scalability, and the privacy of prosumers. Studies like those by Long et al. \cite{long2018peer} and Okwuibe et al. Qiu et al. \cite{qiu2022mean} and Charbonnier et al. \cite{charbonnier2022scalable} proposed scalable solutions for P2P energy trading systems. Studies by Qiu et al. \cite{qiu2021multi} and Lin et al. \cite{lin2022distributed} addressed privacy and security concerns in P2P energy trading.

\subsection{Our Approach}

The proposed models have shown promise in improving efficiency, data privacy, scalability, and cost savings in P2P energy systems. Our research thoroughly examined the literature on P2P energy trading, identifying crucial success factors. Notably, work in this domain-specific to dairy farms has not yet been explored, underscoring the novelty and importance of our study. Our investigation focuses on the unique characteristics of dairy farms, necessitating a specialized MAS-based approach. Our system is customized to optimize energy trading considering dairy farm-specific patterns. By integrating financial benefits, data privacy, scalability, and transparent auction mechanisms, our simulation provides a tailored solution for dairy farms.

This study builds upon our previously proposed simulator outlined in \cite{shah2023multi}, which primarily relied on a rule-based multi-agent system. However, this paper introduces Q-learning into the simulator as a singular agent, alongside the rule-based agents, to evaluate whether the integration of reinforcement learning (RL), particularly Q-learning, enhances the outcomes. While newer RL algorithms offer advancements, Q-learning remains robust and well-established, making it ideal for our research due to its simplicity, interpretability, and effectiveness in addressing our specific problem domain. The results section reveals that the inclusion of a Q-learning trained agent alongside rule-based agents has yielded improvements in key metrics, including reductions in electricity procurement and peak demand, as well as an increase in surplus electricity sales, compared to the solely rule-based simulator. This research contributes to the advancement of:

\begin{itemize}
\item Development of a tailored Q-learning agent for P2P energy trading in dairy farming.
\item Examination of increased revenue, decreased peak demand, and reduced electricity costs achieved through integrating an RL agent within a rule-based agent environment.
\end{itemize}

\section{Background}

\subsection{Rule-based Systems}
The essence of a rule-based system lies in its collection of IF-THEN rules, which encapsulates domain knowledge provided by experts or gleaned from historical data, alongside an inference engine enabling the system to generate outputs based on these rules \cite{yang2022highly}.

A proficient rule-based system holds promise for achieving a favourable balance between accuracy, efficiency, and explainability. Despite widespread utilization of off-the-shelf rule-based systems across diverse domains, challenges persist due to opaque rule relations, efficiency issues in inference, ineffective search strategies, and constraints in knowledge acquisition, limiting their market viability and posing ongoing challenges \cite{furnkranz2020cognitive}.

\subsection{Reinforcement Learning}
Reinforcement learning is a branch of machine learning focused on training agents to make decisions based on rewards and penalties. In RL, agents interact with an environment, receive feedback, and adjust their actions to achieve specific objectives. The goal is to maximize cumulative rewards by learning a policy that dictates actions based on observed states. RL algorithms can be categorized as either value-based like Q-learning algorithm, aiming to learn the optimal value function, or policy-based like REINFORCE and Actor-Critic, aiming to learn the optimal policy directly. RL is instrumental in developing autonomous agents capable of learning from experience, refining decision-making processes, and enhancing their behaviour over time \cite{dadman2023multi}.

\subsection{Q-learning}

Q-learning, a widely adopted reinforcement learning algorithm, is renowned for its simplicity and efficacy in solving Markov decision processes (MDPs) without requiring a model of the environment \cite{watkins1992q}. In the realm of P2P energy trading, Q-learning offers a promising avenue for optimizing energy transactions and resource allocation among distributed participants. By leveraging past experiences and rewards, agents can learn optimal strategies for buying, selling, and storing energy, thereby enhancing the efficient utilization of renewable energy resources and grid infrastructure \cite{shah2024multi}. Through exploration and exploitation, agents adapt to dynamic market conditions, maximizing energy trading profits while minimizing reliance on centralized utility grids. Advanced variants of Q-learning, such as Deep Q-Networks (DQN) and Prioritized Experience Replay, further bolster its effectiveness and scalability in P2P energy trading scenarios \cite{mnih2015human}. By harnessing Q-learning and its derivatives, stakeholders in the energy sector can develop intelligent systems that optimize energy management, reduce costs, and promote sustainability in decentralized energy markets.

In Q-learning, action-value function $Q(s,a)$, represents the expected cumulative reward for taking actions $a$ iteratively in state $s$ and following the optimal policy thereafter. The update equation is as follows:

\[ Q(s,a) \leftarrow (1 - \alpha) \cdot Q(s,a) + \alpha \cdot \left( r + \gamma \cdot \max_{a'} Q(s',a') \right) \]

Here, $\alpha$ denotes the learning rate, $r$ signifies the immediate reward, $\gamma$ represents the discount factor, and $s'$ denotes the next state. The policy for action selection, often $\epsilon$-greedy, balances exploration and exploitation.

\section{System Design}

\begin{figure*}[h!]
    \centering
    \includegraphics[width=1\textwidth]{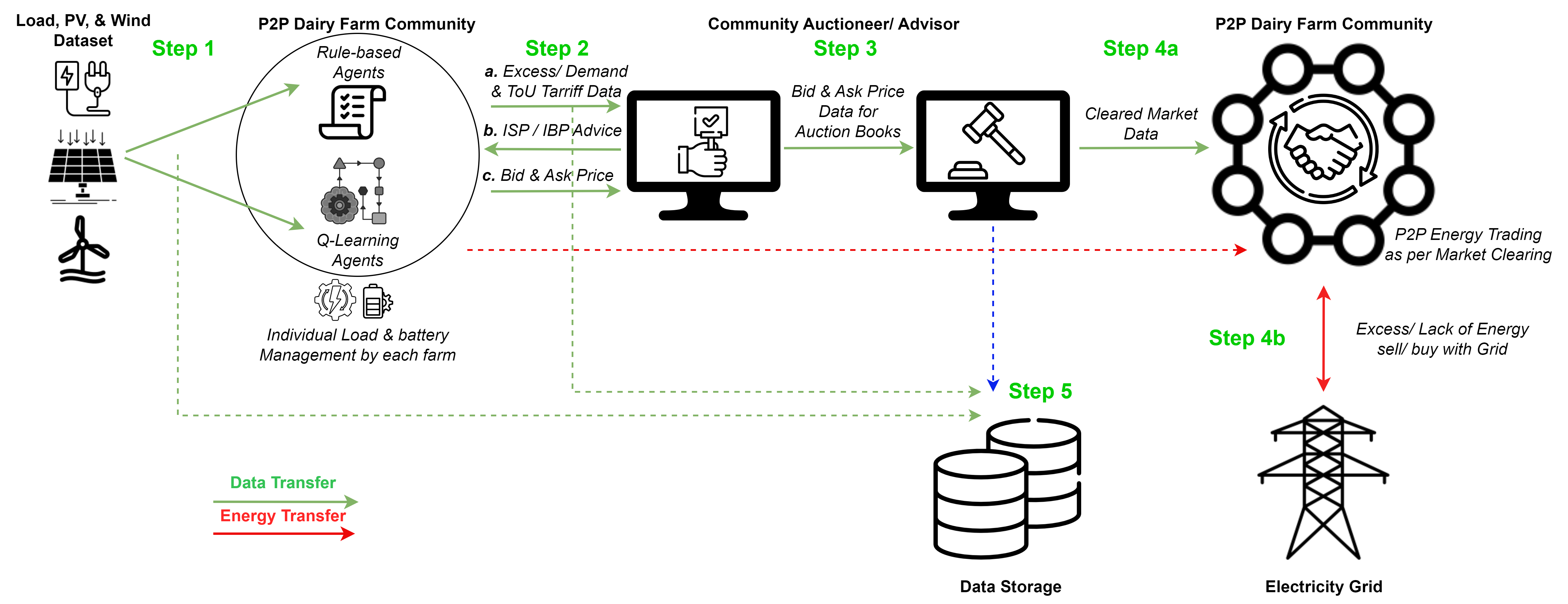}
    \caption{Process Flow for MAPDES P2P Energy Trading Simulator}
    \label{methodology figure}
\end{figure*}

The research aims to present a MAS algorithm incorporating both rule-based and Q-learning RL agents to facilitate distributed P2P energy trading among dairy farms. The Q-learning agent is trained independently in its environment. Once the Q-table is developed, it is integrated into the pre-built rule-based simulator. By employing MAS, the proposed approach enables dairy farms to trade surplus renewable energy resources with neighbouring farms, thereby reducing their dependence on the utility grid and fostering energy self-sufficiency. Figure \ref{methodology figure} depicts the process flow of the simulation model, which is elaborated upon in subsequent sections outlining each step in detail.

\subsection{Dataset and Infrastructure}

The study presents a Python-based simulation model for computing energy generation, consumption, and storage on farms using RE like solar and wind. The model allows for battery storage and enables energy trading within a P2P network or with the grid. The farm load data originates from Khaleghy et al.'s dairy farm load modelling algorithm \cite{khaleghy2023modelling}. PV and wind generation data, varying with farm size, are sourced from SAM for County Dublin, Ireland, and jgmill et al. \cite{samNrel},\cite{windireland}. PV system capacities range from 10 kW to 20 kW, with a fixed 10 kW turbine capacity for wind power. Tesla PowerWall features are considered for farms with battery storage \cite{Teslapower}.

The infrastructure supports scalable farm numbers, each with PV, wind systems, and battery storage. All farms act as prosumers, connected to the grid for energy exchange based on Internal Selling Price (ISP) and Internal Buying Price (IBP), determined by Supply Demand Ratio (SDR) hourly. The simulator encompasses load and battery management, ISP/IBP calculation, and market clearing through auction and energy trading, adaptable for various durations \cite{khaleghy2023modelling}.

A central auctioneer/advisor sets ISP and IBP using SDR hourly, facilitating fair energy exchange within the P2P network. The infrastructure operates on a distributed peer-to-peer model, with market clearance controlled by Double Auction (DA) algorithm. Participants independently manage their load, generation, and battery usage without disclosing data, ensuring efficient energy distribution. Agents include farms, batteries, renewable energy sources, auctioneers, and energy traders, collaborating for effective energy management. Load and battery management is divided into rule-based and Q-learning-based approaches.

\subsection{Rule-Based Training}

The rule-based load and battery management, along with all the related algorithms and equations, have already been discussed in detail in our previous work \cite{shah2023multi}. However, a summarized discussion has been included in this paper.

Input parameters include energy consumption, renewable source capacity, and battery storage. Energy prices are determined based on time of day, renewable source generation, and initial battery storage. The model calculates farm energy consumption and checks if renewable resources can meet the demand. If not, the farm may purchase energy from the grid or other farms via a P2P network. The model simulates various variables such as total energy generation, RE resource availability, battery storage status, and energy prices to manage energy generation, consumption, and storage effectively. It also considers scenarios like RE resource availability and battery storage state to determine energy trading with the grid or other farms. The electricity price for purchasing from the grid is categorized into night, day, and peak rates, depending on the location of the farms. The model includes equations for battery management, energy buy/sell decisions, and electricity pricing based on the Supply Demand Ratio (SDR). Market clearing and energy trading are facilitated using a Double Auction (DA) algorithm. The DA market efficiently matches multiple buyers and sellers for energy trading. It is a widely used mechanism for trading various commodities, such as stocks and electricity. The DA market operates during a fixed auction period, with this research using an hourly resolution. Traders submit their bids/offers at the start of the auction period based on the ISP and IBP, in a controlled manner. The auctioneer then clears the market and publishes the outcomes (trading prices and quantities) at the end of each period \cite{qiu2022mean}. This ensures efficient energy distribution and fair market outcomes.

\subsection{Q-learning based Training}

\begin{figure}[h!]
    \centering
    \includegraphics[width=0.7\textwidth]{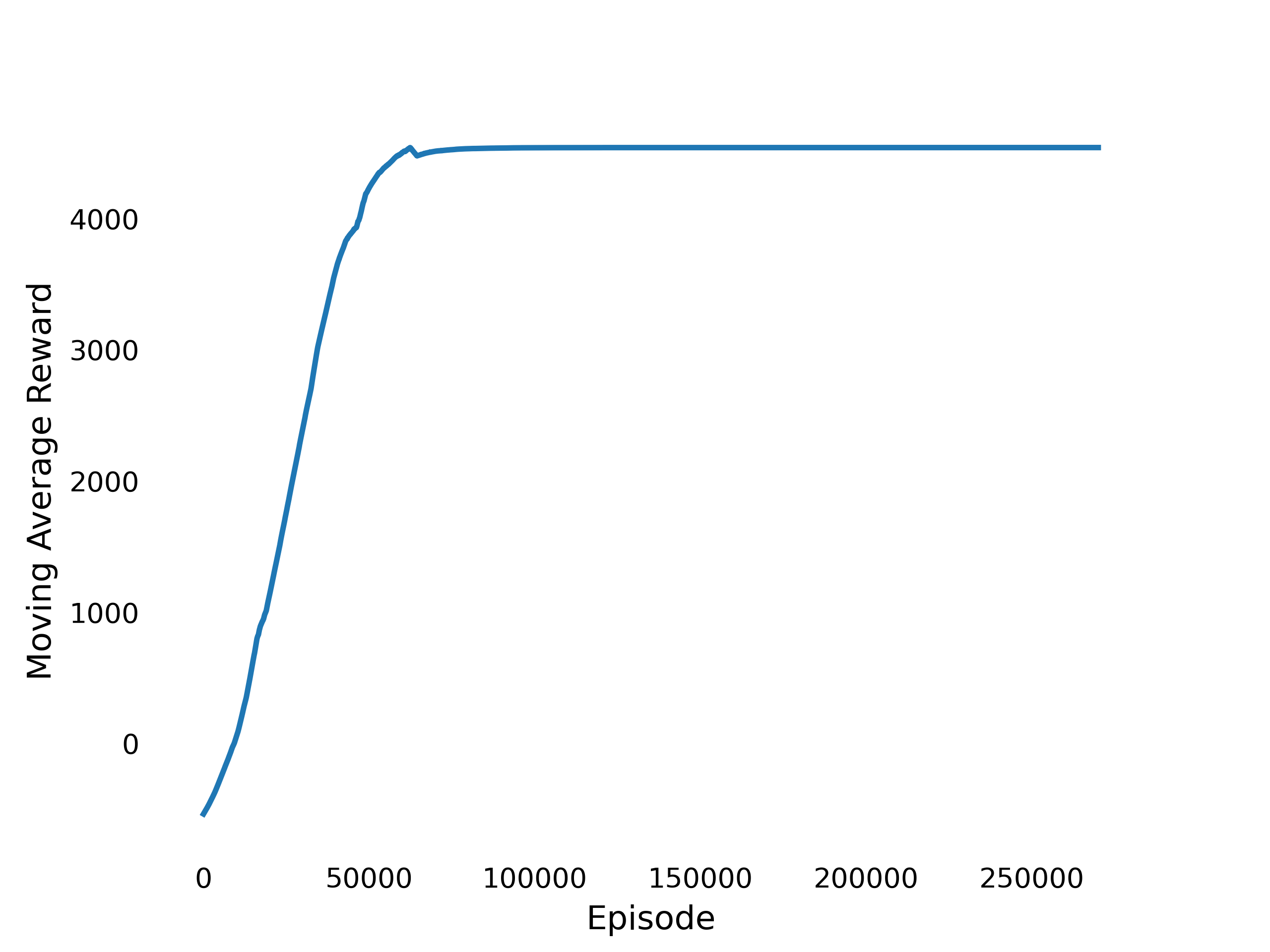}
    \caption{Q-learning Algorithm Convergence}
    \label{convergence}
\end{figure}

The Q-learning algorithm employed in this study underwent training over 300,000 episodes with specific parameters: a learning rate of 0.1, a discount factor of 0.99, an exploration rate starting at 1.0 and decaying by 0.99. The observation space encompasses four features: the farm's load profile, renewable energy generation profile, battery percentage, and the current hour. The action space comprised nine actions: buy, sell, self-consumption only, battery charge and sell, charge and buy, battery discharge and sell, discharge and buy, self-utilization and charge, and self-utilization and discharge. Rewards were determined based on electricity cost from grid tariff, RE generation, farm load, and peak hours. Correct actions yielded positive rewards corresponding to the amount of electricity sold or bought, while incorrect actions led to negative rewards. The rewards converged at around the 60000th episode, as seen in Figure \ref{convergence}. The convergence line shown in this figure is an average value line of rewards per episode with a window size of 200 (used to smooth the learning curve by averaging recent experiences). 

\section{Results and Discussions}

We evaluated our approach by simulating 10 dairy farms over one year, comprising one Q-learning-based agent and nine rule-based agents. Using hourly time steps, we assessed the simulation results based on four key metrics:

\begin{enumerate}
    \item[$(i)$] Comparison of energy purchased by farms without RE resources and P2P vs with RE resources and P2P.
    \item[$(ii)$] Electricity purchase among RE-equipped farms (P2P vs Non-P2P).
    \item[$(iii)$] Electricity sold by RE-equipped farms (P2P vs non-P2P).
    \item[$(iv)$] Peak hours grid demand by RE-equipped farms (P2P vs non-P2P).
\end{enumerate}

\begin{figure}[h!]
    \centering
    \includegraphics[width=0.7\textwidth]{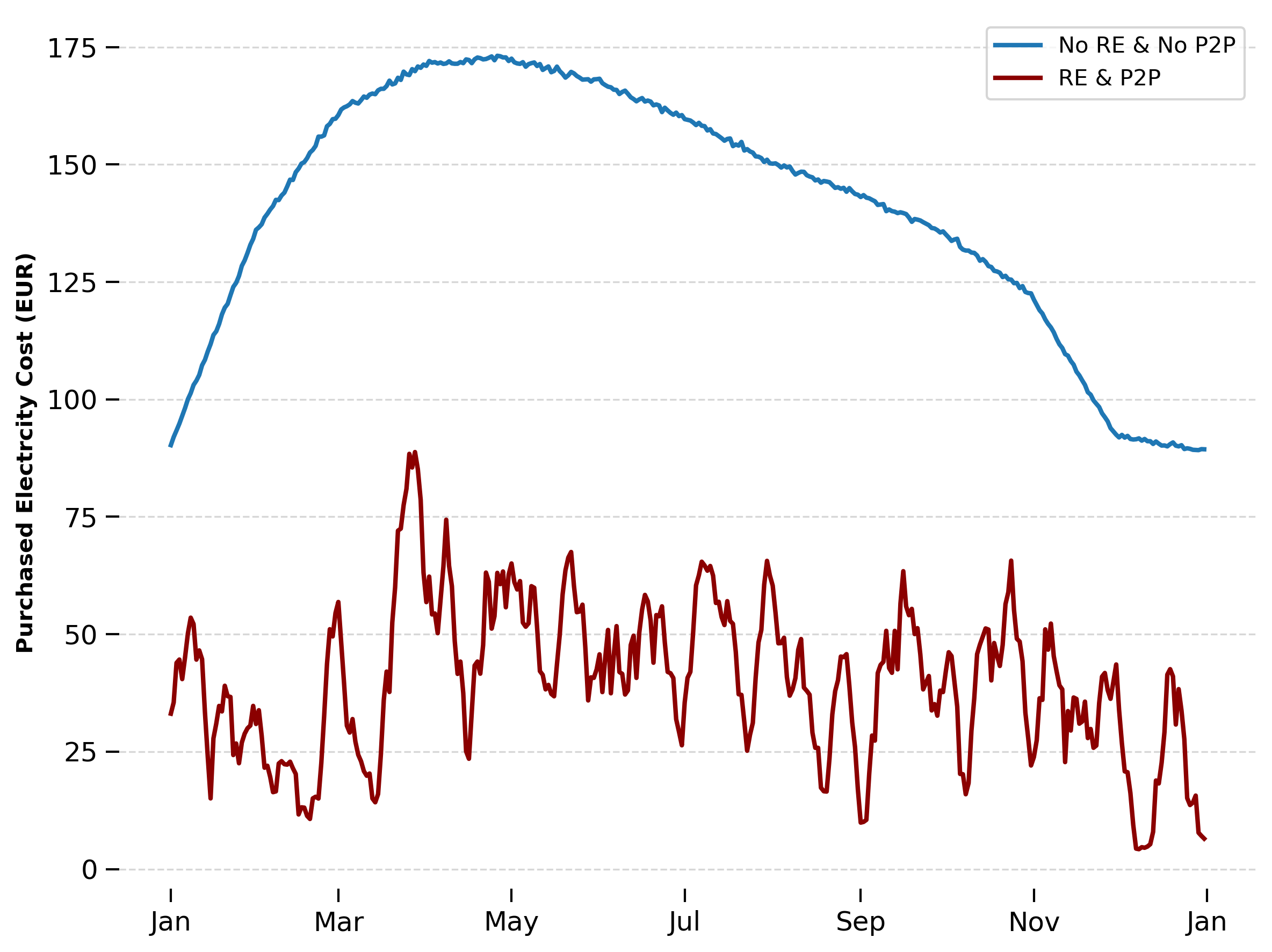}
    \caption{Total Electricity Cost: P2P \& RE vs Non-P2P \& No RE}
    \label{boughtwoRE}
\end{figure}

Figure \ref{boughtwoRE} illustrates the contrast in total energy procurement between farm communities employing RE resources and engaging in P2P energy trading, versus those without RE resources and P2P trading. Notably, there's a 70.44\% reduction in electricity purchases for a farm community when RE resources and P2P trading are utilized.

\begin{figure}[h!]
    \centering
    \includegraphics[width=0.7\textwidth]{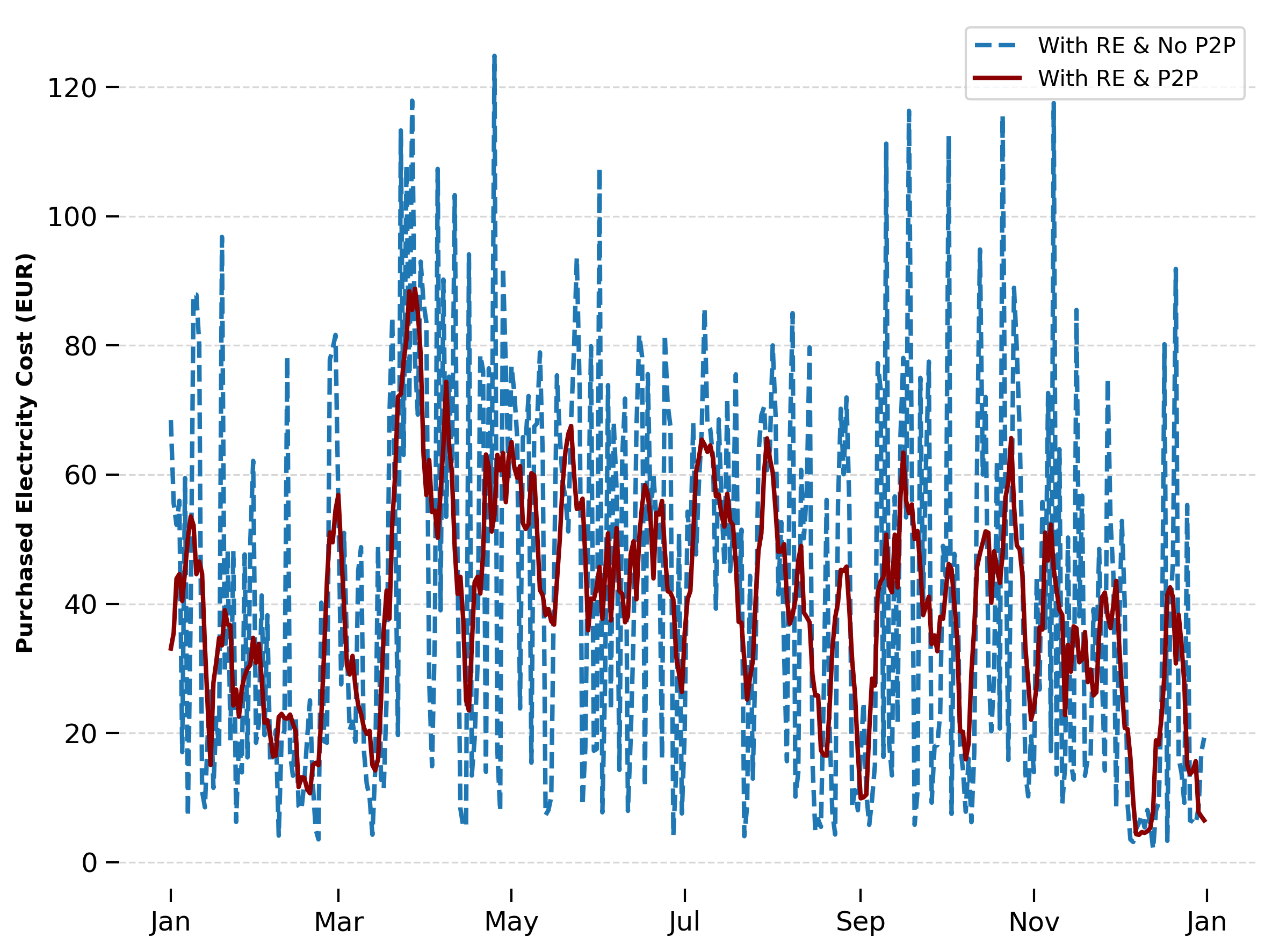}
    \caption{Total Electricity Cost: P2P vs Non-P2P (RE-equipped)}
    \label{boughtwithRE}
\end{figure}

\begin{figure}[h!]
    \centering
    \includegraphics[width=0.7\textwidth]{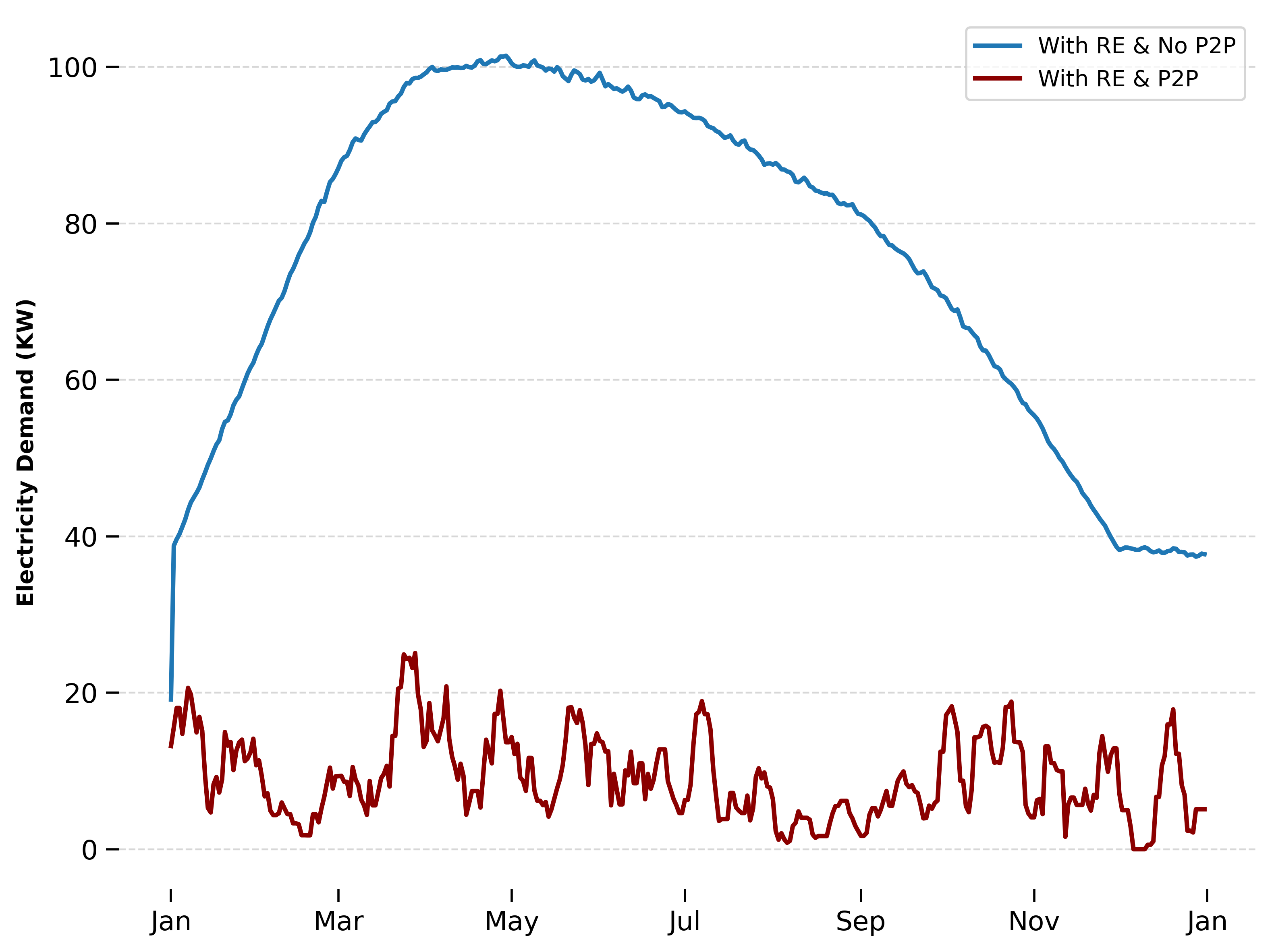}
    \caption{Electricity demand of Farm Community from the grid in peak hours (5 pm to 7 pm): P2P vs Non-P2P}
    \label{peakdemand}
\end{figure}

Furthermore, Figure \ref{boughtwithRE} compares total energy purchases between farm communities, all using RE resources, but with and without P2P energy trading. Here, we observe a 4.13\% decrease in electricity procurement with P2P trading, despite both scenarios having RE resources. While this percentage reduction may seem modest, it's essential to consider the significant impact during peak hours, as depicted in Figure \ref{peakdemand}. P2P energy trading notably reduces grid reliance during peak hours by a remarkable 87.84\%.

\begin{figure}[h!]
    \centering
    \includegraphics[width=0.7\textwidth]{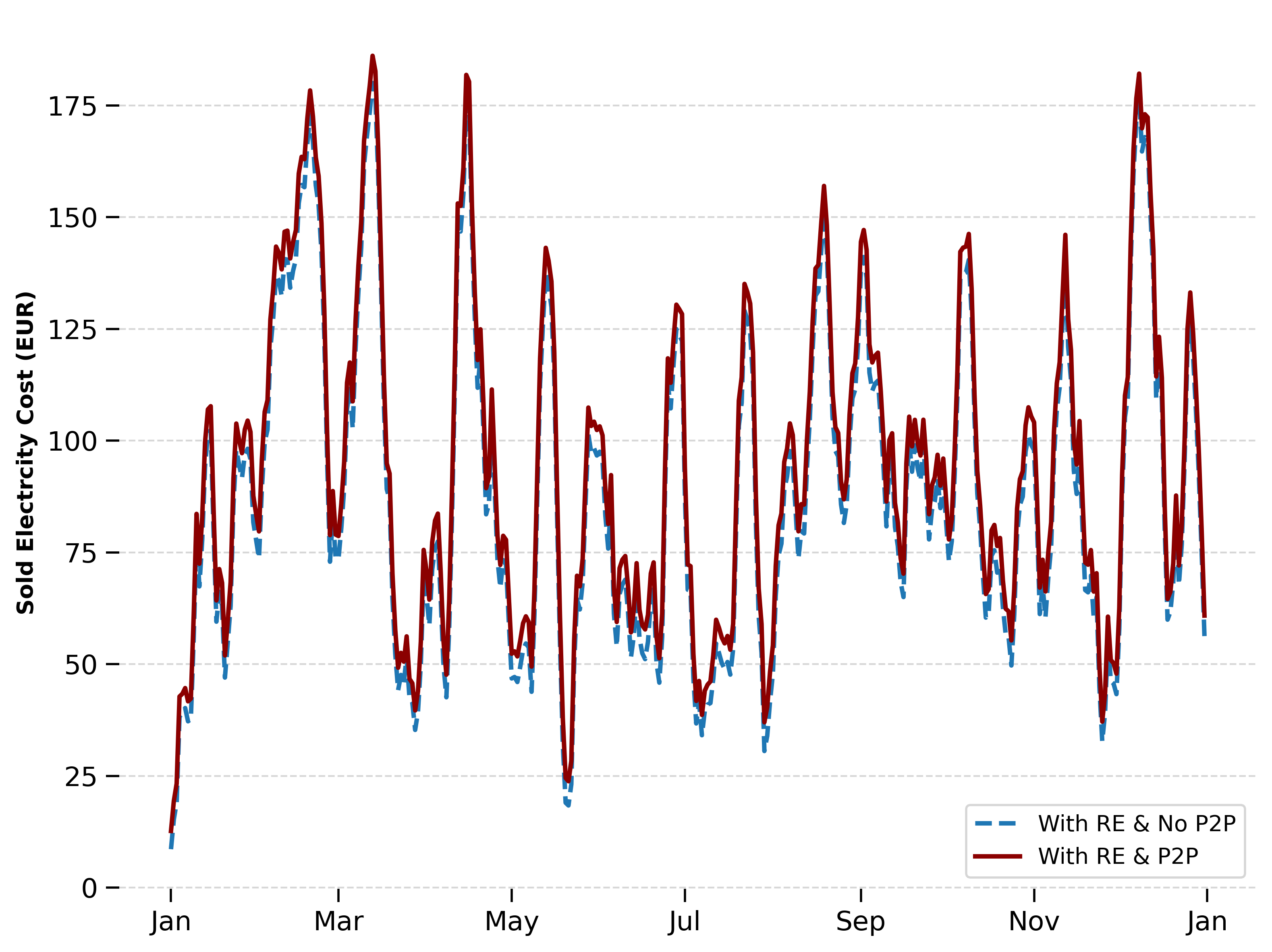}
    \caption{Total Daily Revenue from Excess Electricity Sold using P2P vs non-P2P}
    \label{EnergySold}
\end{figure}

Moreover, the adoption of P2P energy trading leads to a 1.91\% increase in revenue generated from electricity sales, as demonstrated in Figure \ref{EnergySold}.

When comparing the performance metrics between our approach, which integrates Q-learning-based agents with rule-based agents, and the results from a purely rule-based approach, our method shows promising outcomes. Specifically, our approach yields lower electricity purchase costs, reduced peak demand, and enhanced revenue from electricity sales, as illustrated in Table \ref{ComparisionTable}. This indicates that Q-learning outperforms the rule-based approach in dairy farming P2P energy trading.

\begin{table}[!ht]
\centering
\caption{MAPDES Results for Different Locations: Finland vs Ireland}
\label{ComparisionTable}
\small 
\begin{tabular}{p{8cm}p{2cm}p{2cm}}

\hline

 & \textbf{Rule-based} & \textbf{Combined (Q-learning \& Ruled)} \\
\hline
\textbf{Electricity Cost with no RE (€)} & 51710 & 51710 \\

\textbf{Electricity Cost with RE no P2P (€)} & 21066 & 15284 \\

\textbf{Electricity Cost with P2P and RE (€)} & 20497 & 14653 \\

\textbf{Electricity Cost \% reduction with P2P and RE vs w/o RE (\%)} & \textbf{59.26} & \textbf{70.44} \\

\textbf{Electricity Cost \% reduction with P2P and RE vs with RE only (\%)} & \textbf{2.69} & \textbf{4.13} \\

\textbf{Electricity Revenue without P2P (€)} & 46032 & 46022 \\

\textbf{Electricity Revenue with P2P (€)} & 46612 & 46904 \\

\textbf{Electricity Revenue \% increase P2P vs no P2P (\%)} & \textbf{1.25} & \textbf{1.91} \\

\textbf{Peak Demand without P2P (kW)} & 27837 & 27837 \\

\textbf{Peak Demand with P2P (kW)} & 8618 & 3382 \\

\textbf{peak Demand \% reduction using P2P vs no P2P (\%)} & \textbf{69.03} & \textbf{87.84} \\
\hline

\end{tabular}
\end{table}

\section{Conclusion}

This paper introduces a comprehensive framework designed to optimize energy management within dairy farming communities by integrating Q-learning-based agents with rule-based systems. Through extensive simulations and analysis, we have showcased the effectiveness of our approach, achieving remarkable reductions in electricity purchase costs (by 70.44\%), substantial minimization of peak demand (by 87.84\%), and notable maximization of revenue from electricity sales (by 1.91\%). By combining reinforcement learning with traditional rule-based strategies, our method presents a versatile and efficient solution for sustainable energy management in agricultural contexts. Furthermore, our findings underscore the significant advantages of peer-to-peer energy trading, highlighting its potential to revolutionize the dairy farming sector. Looking ahead, future research avenues may explore enhancements to the algorithm, including the utilization of a purely reinforcement learning-based multi-agent system, and extend its application to other agricultural domains, thereby fostering more resilient and eco-friendly farming practices.

\begin{credits}
\subsubsection{\ackname} This publication has emanated from research conducted with the financial support of Science Foundation Ireland under Grant number [21/FFP-A/9040].

\subsubsection{\discintname}
Author has received a research grant from Science Foundation Ireland. 
\end{credits}
%
%

\bibliography{Bibliography}

\begin{thebibliography}{10}
\providecommand{\url}[1]{\texttt{#1}}
\providecommand{\urlprefix}{URL }
\providecommand{\doi}[1]{https://doi.org/#1}

\bibitem{adjerid2020multi}
Adjerid, H., Maouche, A.R.: Multi-agent system-based decentralized state estimation method for active distribution networks. Computers \& Electrical Engineering  \textbf{86},  106652 (2020)

\bibitem{charbonnier2022scalable}
Charbonnier, F., Morstyn, T., McCulloch, M.D.: Scalable multi-agent reinforcement learning for distributed control of residential energy flexibility. Applied Energy  \textbf{314},  118825 (2022)

\bibitem{eurcomclimate}
Commission, E.: Climate change: what the eu is doing. \url{https://www.consilium.europa.eu/en/policies/climate-change/\#2050/} (2024), [Accessed on March 15, 2024]

\bibitem{dadman2023multi}
Dadman, S., Bremdal, B.A.: Multi-agent reinforcement learning for structured symbolic music generation. In: International Conference on Practical Applications of Agents and Multi-Agent Systems. pp. 52--63. Springer (2023)

\bibitem{elena2022multi}
Elena, D.O., Florin, D., Valentin, G., Marius, P., Octavian, D., Catalin, D.: Multi-agent system for smart grids with produced energy from photovoltaic energy sources. In: 2022 14th International Conference on Electronics, Computers and Artificial Intelligence (ECAI). pp.~1--6. IEEE (2022)

\bibitem{elkazaz2021hierarchical}
Elkazaz, M., Sumner, M., Thomas, D.: A hierarchical and decentralized energy management system for peer-to-peer energy trading. Applied Energy  \textbf{291},  116766 (2021)

\bibitem{windireland}
Fraukewiese, sjpfenninger, jgmill: Load, wind and solar, prices in hourly resolution. \url{https://data.open-power-system-data.org/time\_series/latest/} (2020), accessed on January 15, 2024

\bibitem{furnkranz2020cognitive}
F{\"u}rnkranz, J., Kliegr, T., Paulheim, H.: On cognitive preferences and the plausibility of rule-based models. Machine Learning  \textbf{109}(4),  853--898 (2020)

\bibitem{guimaraes2021agent}
Guimar{\~a}es, D.V., Gough, M.B., Santos, S.F., Reis, I.F., Home-Ortiz, J.M., Catal{\~a}o, J.P.: Agent-based modeling of peer-to-peer energy trading in a smart grid environment. In: 2021 IEEE International Conference on Environment and Electrical Engineering and 2021 IEEE Industrial and Commercial Power Systems Europe (EEEIC/I\&CPS Europe). pp.~1--6. IEEE (2021)

\bibitem{khaleghy2023modelling}
Khaleghy, H., Wahid, A., Clifford, E., Mason, K.: Modelling electricity consumption in irish dairy farms using agent-based modelling. arXiv preprint arXiv:2308.09488  (2023)

\bibitem{khorasany2018market}
Khorasany, M., Mishra, Y., Ledwich, G.: Market framework for local energy trading: A review of potential designs and market clearing approaches. IET Generation, Transmission \& Distribution  \textbf{12}(22),  5899--5908 (2018)

\bibitem{khorasany2019decentralized}
Khorasany, M., Mishra, Y., Ledwich, G.: A decentralized bilateral energy trading system for peer-to-peer electricity markets. IEEE Transactions on industrial Electronics  \textbf{67}(6),  4646--4657 (2019)

\bibitem{lin2022distributed}
Lin, W.T., Chen, G., Zhou, X.: Distributed carbon-aware energy trading of virtual power plant under denial of service attacks: A passivity-based neurodynamic approach. Energy  \textbf{257},  124751 (2022)

\bibitem{long2018peer}
Long, C., Wu, J., Zhou, Y., Jenkins, N.: Peer-to-peer energy sharing through a two-stage aggregated battery control in a community microgrid. Applied energy  \textbf{226},  261--276 (2018)

\bibitem{mnih2015human}
Mnih, V., Kavukcuoglu, K., Silver, D., Rusu, A.A., Veness, J., Bellemare, M.G., Graves, A., Riedmiller, M., Fidjeland, A.K., Ostrovski, G., et~al.: Human-level control through deep reinforcement learning. nature  \textbf{518}(7540),  529--533 (2015)

\bibitem{samNrel}
{National Renewable Energy Lab (NREL)}: System advisor model (sam). \url{https://sam.nrel.gov} (2017), version 2017.9.5

\bibitem{qiu2022mean}
Qiu, D., Wang, J., Dong, Z., Wang, Y., Strbac, G.: Mean-field multi-agent reinforcement learning for peer-to-peer multi-energy trading. IEEE Transactions on Power Systems  (2022)

\bibitem{qiu2021multi}
Qiu, D., Wang, J., Wang, J., Strbac, G.: Multi-agent reinforcement learning for automated peer-to-peer energy trading in double-side auction market. In: IJCAI. pp. 2913--2920 (2021)

\bibitem{qiu2021scalable}
Qiu, D., Ye, Y., Papadaskalopoulos, D., Strbac, G.: Scalable coordinated management of peer-to-peer energy trading: A multi-cluster deep reinforcement learning approach. Applied Energy  \textbf{292},  116940 (2021)

\bibitem{shah2023multi}
Shah, M.I.A., Wahid, A., Barrett, E., Mason, K.: A multi-agent systems approach for peer-to-peer energy trading in dairy farming. arXiv preprint:2310.05932  (2023)

\bibitem{shah2024multi}
Shah, M.I.A., Wahid, A., Barrett, E., Mason, K.: Multi-agent systems in peer-to-peer energy trading: A comprehensive survey. Engineering Applications of Artificial Intelligence  \textbf{132},  107847 (2024)

\bibitem{Teslapower}
Tesla: How powerwall works. \url{https://www.tesla.com/support/energy/ powerwall/learn/how-powerwall-works} (2023), [Online; Accessed March 20, 2023]

\bibitem{umer2021novel}
Umer, K., Huang, Q., Khorasany, M., Afzal, M., Amin, W.: A novel communication efficient peer-to-peer energy trading scheme for enhanced privacy in microgrids. Applied Energy  \textbf{296},  117075 (2021)

\bibitem{upton2015investment}
Upton, J., Murphy, M., De~Boer, I., Koerkamp, P.G., Berentsen, P., Shalloo, L.: Investment appraisal of technology innovations on dairy farm electricity consumption. Journal of dairy science  \textbf{98}(2),  898--909 (2015)

\bibitem{upton2013energy}
Upton, J., Humphreys, J., Koerkamp, P.G., French, P., Dillon, P., De~Boer, I.J.: Energy demand on dairy farms in ireland. Journal of dairy science  \textbf{96}(10),  6489--6498 (2013)

\bibitem{watkins1992q}
Watkins, C.J., Dayan, P.: Q-learning. Machine learning  \textbf{8},  279--292 (1992)

\bibitem{yang2022three}
Yang, J., Xu, W., Ma, K., Li, C.: A three-stage multi-energy trading strategy based on p2p trading mode. IEEE Transactions on Sustainable Energy  \textbf{14}(1),  233--241 (2022)

\bibitem{yang2022highly}
Yang, L.H., Liu, J., Ye, F.F., Wang, Y.M., Nugent, C., Wang, H., Mart{\'\i}nez, L.: Highly explainable cumulative belief rule-based system with effective rule-base modeling and inference scheme. Knowledge-Based Systems  \textbf{240},  107805 (2022)

\bibitem{zhang2019p2p}
Zhang, M., Eliassen, F., Taherkordi, A., Jacobsen, H.A., Chung, H.M., Zhang, Y.: Energy trading with demand response in a community-based p2p energy market. In: 2019 IEEE International Conference on Communications, Control, and Computing Technologies for Smart Grids (SmartGridComm). pp.~1--6 (2019). \doi{10.1109/SmartGridComm.2019.8909798}

\bibitem{zhou2020decentralized}
Zhou, H., Erol-Kantarci, M.: Decentralized microgrid energy management: A multi-agent correlated q-learning approach. In: 2020 IEEE International Conference on Communications, Control, and Computing Technologies for Smart Grids (SmartGridComm). pp.~1--7. IEEE (2020)

\bibitem{zhou2018evaluation}
Zhou, Y., Wu, J., Long, C.: Evaluation of peer-to-peer energy sharing mechanisms based on a multiagent simulation framework. Applied energy  \textbf{222},  993--1022 (2018)

\bibitem{zhu2022market}
Zhu, L.: Market-based versus price-based optimal trading mechanism design in microgrid. Computers and Electrical Engineering  \textbf{100},  107904 (2022)

\end{thebibliography}

\end{document}